%% file: emnlp2022.tex
\documentclass[11pt]{article}

\usepackage{EMNLP2022}

\usepackage{times}
\usepackage{latexsym}

\usepackage[T1]{fontenc}

\usepackage[utf8]{inputenc}

\usepackage{microtype}

\usepackage{inconsolata}

\usepackage{amsmath}
\usepackage{amssymb}
\usepackage{subcaption}
\usepackage{graphicx}
\usepackage{booktabs}
\usepackage{multirow}
\usepackage{soul}
\usepackage{float}
\newcommand*\samethanks[1][\value{footnote}]{\footnotemark[#1]}

\title{COST-EFF: Collaborative Optimization of Spatial and Temporal Efficiency with Slenderized Multi-exit Language Models}

\author{
  Bowen Shen$^{1,2}$, Zheng Lin$^{1,2}$\Thanks{\ \ Zheng Lin and Lei Wang are the corresponding authors.}\ , Yuanxin Liu$^{1,3}$, Zhengxiao Liu$^{1}$, \\
  \bf Lei Wang$^{1}$\samethanks[1]\ , Weiping Wang$^{1}$ \\
  $^{1}$Institute of Information Engineering, Chinese Academy of Sciences\\
  $^{2}$School of Cyber Security, University of Chinese Academy of Sciences\\
  $^{3}$MOE Key Laboratory of Computational Linguistics, Peking University \\
  \texttt{\{shenbowen, linzheng, liuzhengxiao, wanglei, wangweiping\}@iie.ac.cn}\\
  \texttt{liuyuanxin@stu.pku.edu.cn}
}

\begin{document}
\maketitle
\begin{abstract}
  \input{abstract.tex}
\end{abstract}

\input{body.tex}

\section*{Limitations}

COST-EFF currently has the following limitations. If they are addressed in future works, the potential capabilities of COST-EFF can be unleashed.
(1) During the inference of dynamic early exiting models, the conventional practice is to set batch size as 1 to better adjust the computational according to individual input samples. However, such a setting is not always effective as a larger batch size is likely to reduce inference time, whereas input complexities inside a batch may differ significantly. Thus, it is inspiring to investigate a pipeline that gathers samples with similar expectations of complexity into a batch while controlling the priority of batches with different complexities to achieve parallelism.
(2) We choose natural language understanding (NLU) tasks to study compression and acceleration following the strong baselines TinyBERT \cite{jiao2020tinybert} and ElasticBERT \cite{liu_towards_2022}. However, the extensibility of COST-EFF is yet to be explored in more complex tasks including natural language generation, translation, etc. So far, static model compression is proved to be effective in complex tasks \cite{gupta2022compression} and we are seeking the extension of dynamic inference acceleration on different tasks using models with an iterative process.

\section*{Acknowledgements}

This work was supported by National Natural Science Foundation of China (No. 61976207, No. 61906187).

\bibliography{anthology,custom}
\bibliographystyle{acl_natbib}

\vfill\break

\appendix

\input{appendix.tex}


\end{document}

%% file: abstract.tex
Transformer-based pre-trained language models (PLMs) mostly suffer from excessive overhead despite their advanced capacity.
For resource-constrained devices, there is an urgent need for a spatially and temporally efficient model which retains the major capacity of PLMs.
However, existing statically compressed models are unaware of the diverse complexities between input instances, potentially resulting in redundancy and inadequacy for simple and complex inputs. Also, miniature models with early exiting encounter challenges in the trade-off between making predictions and serving the deeper layers.
Motivated by such considerations, we propose a collaborative optimization for PLMs that integrates static model compression and dynamic inference acceleration.
Specifically, the PLM is slenderized in width while the depth remains intact, complementing layer-wise early exiting to speed up inference dynamically.
To address the trade-off of early exiting, we propose a joint training approach that calibrates slenderization and preserves contributive structures to each exit instead of only the final layer.
Experiments are conducted on GLUE benchmark and the results verify the Pareto optimality of our approach at high compression and acceleration rate with 1/8 parameters and 1/19 FLOPs of BERT.

%% file: body.tex
\section{Introduction}

\begin{figure}[t]
    \centering
    \includegraphics[width=0.9\columnwidth]{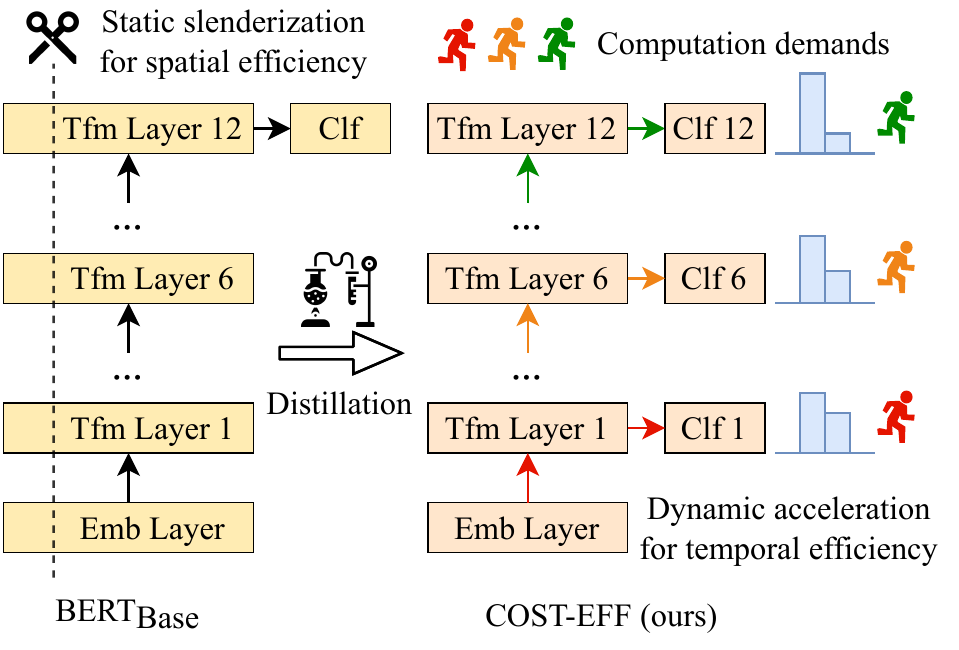}
    \caption{An illustration of COST-EFF model structure and inference procedure. Emb, Tfm and Clf are abbreviations of embedding, Transformer and classifier, respectively. Blue bar charts denote probability distribution output by classifiers.}
    \label{fig:intro}
\end{figure}

Pre-training generalized language models and fine-tuning them on specific downstream tasks has become the dominant paradigm in natural language processing (NLP) since the advent of Transformers \cite{vaswani2017attention} and BERT \cite{devlin2019bert}.
However, pre-trained language models (PLMs) are predominantly designed to be vast in the pursuit of model capacity and generalization. With such a concern, the model storage and inference time of PLMs are usually high, making them challenging to be deployed on resource-constrained devices \cite{sun2020mobilebert}.

Recent studies indicate that Transformer-based PLMs bear redundancy spatially and temporally which comes from the excessive width and depth of the model \cite{michel2019sixteen,xin2021berxit}.
With static compression methods including network pruning \cite{xia2022structured} and knowledge distillation \cite{jiao2020tinybert}, spatial overheads of PLMs (i.e., model parameters) can be reduced to a fixed setting. From the perspective of input instances rather than the model, early exiting without passing all the model layers enables the dynamic acceleration at inference time and diminishes the temporal overheads \cite{zhou2020bert}.

However, static compression can hardly find an optimal setting that is both efficient on simple input instances and accurate on complex ones, while early exiting cannot diminish the redundancy in model width and is impotent for reducing the actual volume of model.
Further, interpretability studies indicate that the attention and semantic features across layers are different in BERT \cite{clark2019bert}.
Hence, deriving a multi-exit model from a pre-trained single-exit model like BERT incurs inconsistency in the training objective, where each layer is simultaneously making predictions and serving the deeper layers \cite{xin2021berxit}. Empirically, we find that the uncompressed BERT is not severely influenced by such inconsistency, whereas small capacity models are not capable of balancing shallow and deep layers. Plugging in exits after compression will lead to severe performance degradation, which hinders the complementation of the two optimizations.

To fully exploit the efficiency of PLMs and mitigate the above-mentioned issues, we design a slenderized multi-exit model and propose a Collaborative Optimization approach of Spatial and Temporal EFFiciency (COST-EFF) as depicted in \autoref{fig:intro}. Unlike previous works, e.g., DynaBERT \cite{hou2020dynabert} and CoFi \cite{xia2022structured}, which obtain a squat model, we keep the depth intact while slenderizing the PLM.
Superiority of slender architectures over squat ones is supported by \cite{bengio2007scaling} and \cite{turc2019well} in generic machine learning and PLM design.
To address the inconsistency in compressed multi-exit model, we first distill an multi-exit BERT from the original PLM as both the teaching assistant (TA) and the slenderization backbone, which is more effective in balancing the trade-off between layers than compressed models.
Then, we propose a collaborative approach that slenderizes the backbone with the calibration of exits.
Such a slenderization diminishes less contributive structures to each exit as well as the redundancy in width.
After the slenderization, task-specific knowledge distillation is conducted with the objectives of hidden representations and predictions of each layer as recovery.
Specifically, the contributions of this paper are as follows.

\begin{itemize}
    \item To comprehensively optimize the spatial and temporal efficiency of PLMs, we leverage both static slenderization and dynamic acceleration from the perspective of model scale and variable computation.

    \item We propose a collaborative training approach that calibrates the slenderization under the guidance of intermediate exits and mitigates the inconsistency of early exiting.

    \item Experiments conducted on the GLUE benchmark verify the Pareto optimality of our approach. COST-EFF achieves 96.5\% performance of fine-tuned BERT$_\text{Base}$ with approximately $1/8$ parameters and $1/19$ FLOPs without any form of data augmentation.\footnote{Code is available at \url{https://github.com/sbwww/COST-EFF}.}
\end{itemize}

\section{Related Work}

The compression and acceleration of PLMs were recently investigated to neutralize the overhead of large models by various means.

The structured pruning objects include, from small to large, hidden dimensions \cite{wang2020structured}, attention heads \cite{michel2019sixteen}, multi-head attention (MHA) and feed-forward network (FFN) modules \cite{xia2022structured} and entire Transformer layers \cite{Fan2020Reducing}. Considering the benefit of the overall structure, we keep all the modules while reducing their sizes. Besides pruning out structures, a fine-grained approach is unstructured pruning which prunes out weights. Unstructured pruning can achieve high sparsity of 97\% \cite{xu2021dense} but is not yet adaptable to general computing platforms and hardware.

During the recovery training of compressed models, knowledge distillation objectives include predictions of classifiers \cite{sanh2019distilbert}, features of intermediate representations \cite{jiao2020tinybert} and relations between samples \cite{tung2019similarity}. Also, the occasion of distillation varies from general pre-training and task-specific fine-tuning \cite{turc2019well}. Distillation enables the training without ground-truth labels complementing data augmentation. In this paper, data augmentation is not leveraged as it requires a long training time, but our approach is well adaptable to it if better performance is to be pursued.

Dynamic early exits come from BranchyNet \cite{teerapittayanon2016branchynet}, which introduces exit branches after specific convolution layers of the CNN model. The idea is adopted to PLMs as Transformer layer-wise early exiting \cite{xin2021berxit,zhou2020bert,liu2020fastbert}. However, early exiting only accelerates inference but does not reduce the model size and the redundancy in width. Furthermore, owing to the inconsistency between shallow and deep layers, it is hard to achieve high speedup using early exiting alone.

The prevailing PLMs, e.g., RoBERTa \cite{liu2019roberta} and XLNet \cite{yang2019xlnet} are variants of Transformer with similar overall structures, well-adaptable to the optimizations that we propose. Apart from PLMs with increasing size, ALBERT\cite{Lan2020ALBERT} is distinctive with a small volume of 18M (Million) parameters obtained from weight sharing of Transformer layers. Weight sharing allows the model to store the parameters only once, greatly reducing the storage overhead. However, the shared weights have no contribution to inference speedup. Instead, the time required for ALBERT to achieve BERT-like accuracy increases.

\section{Methodology}\label{sec:methodology}

In this section, we analyze the major structures of Transformer-based PLMs and devise corresponding optimizations. The proposed COST-EFF has three key properties, namely static slenderization, dynamic acceleration and collaborative training.

\subsection{Preliminaries}\label{sec:preliminaries}

In this paper, we focus on optimizing the Transformer-based PLM which mainly consists of embedding, MHA and FFN. Specifically, embedding converts each input token to a tensor of size $H$ (i.e., hidden dimension). With a common vocabulary size of $\left\lvert \mathbb{V} \right\rvert=30,522$, the word embedding matrix accounts for $< 22\%$ of BERT$_\text{Base}$ parameters. Inside the Transformer, MHA has four matrices $\boldsymbol{W}_Q$, $\boldsymbol{W}_K$, $\boldsymbol{W}_V$ and $\boldsymbol{W}_O$, all of which with input and output size of $H$. FFN has two matrices $\boldsymbol{W}_{FI}$ and $\boldsymbol{W}_{FO}$ with the size of $H \times F$. As the key components of Transformer, MHA and FFN account for $< 26\%$ and $< 52\%$ of BERT$_\text{Base}$ parameters, respectively.

\begin{figure*}[t]
    \centering
    \includegraphics[width=0.9\textwidth]{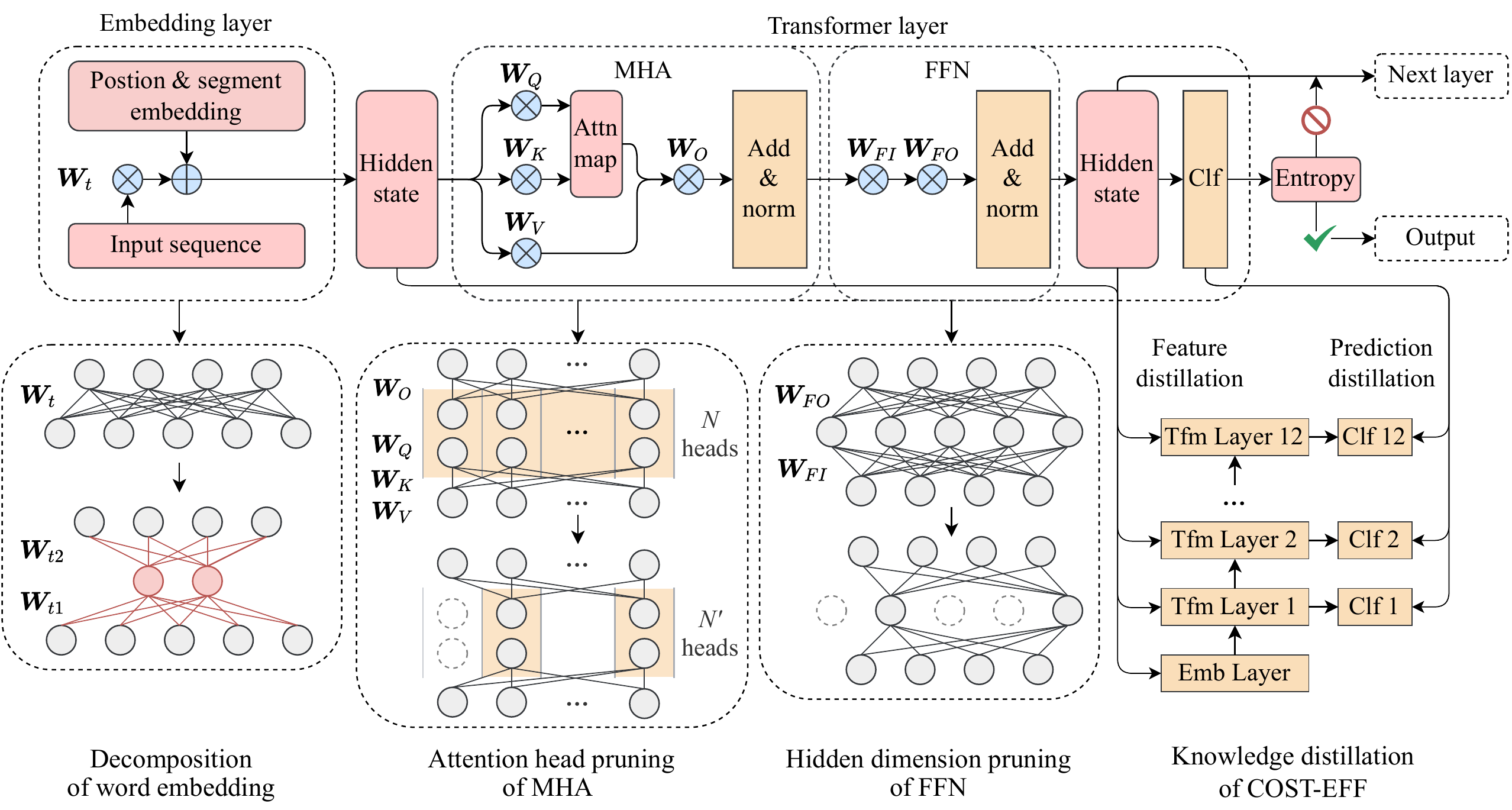}
    \caption{Illustration of COST-EFF. The upper part is the general architecture and forward procedure of the model. The lower part is the slenderization details of corresponding modules, where grey circles denote the input and output dimensions of matrices and the lines connecting them are weights.}
    \label{fig:arch}
\end{figure*}

Based on the analysis, we have the following slenderization and acceleration schemes.
(1) The word embedding matrix $\boldsymbol{W}_t$ is decomposed into the multiplication of two matrices following \cite{Lan2020ALBERT}. Thus, the vocabulary size $\left\lvert \mathbb{V} \right\rvert$ and hidden size $H$ are not changed.
(2) For the transformation matrices of MHA and FFN, structured pruning is adopted to reduce their input or output dimensions.
(3) The inference is accelerated through early exiting as we retain the pre-trained model depth. To avoid introducing additional parameters, we remove the pre-trained pooler matrix before classifiers.
(4) Knowledge distillation on prediction logits and hidden states of each layer is leveraged as a substitute for conventional fine-tuning. The overall architecture of COST-EFF is depicted in \autoref{fig:arch}.

\subsection{Static Slenderization}

\subsubsection{Matrix Decomposition of Embedding}\label{sec:embedding}

As mentioned before, the word embedding takes up more than 1/5 of BERT$_\text{Base}$ parameters.
The output dimension of word embedding is equal to hidden size, which we don't modify, we use truncated singular value decomposition (TSVD) to internally compress the word embedding matrix.

TSVD first decomposes the matrix as $\boldsymbol{A}^{m \times n} = \boldsymbol{U}^{m \times m} \boldsymbol{\Sigma}^{m \times n} \boldsymbol{V}^{n \times n}$, where $\boldsymbol{\Sigma}^{m \times n}$ is the singular value diagonal matrix. After that, the three matrices are truncated to the given rank. Thus, the decomposition of word embedding is as
\begin{equation}
    \begin{aligned}
        \boldsymbol{W}_{t}^{\left\lvert \mathbb{V} \right\rvert \times H} & \approx \boldsymbol{W}_{t1}^{\left\lvert \mathbb{V} \right\rvert \times R} \boldsymbol{W}_{t2}^{R \times H}                                                      \\
                                                                          & =\left(\tilde{\boldsymbol{U}}^{\left\lvert \mathbb{V} \right\rvert \times R} \tilde{\boldsymbol{\Sigma}}^{R \times R}\right) \tilde{\boldsymbol{V}}^{R \times H}
        ,
    \end{aligned}
    \label{eq:svd}
\end{equation}
where we multiplies $\tilde{\boldsymbol{U}}$ and $\tilde{\boldsymbol{\Sigma}}$ matrices as the first embedding matrix $\boldsymbol{W}_{t1}^{\left\lvert \mathbb{V} \right\rvert \times R}$ and $\boldsymbol{W}_{t2}^{R \times H} = \tilde{\boldsymbol{V}}$ is a linear transformation with no bias.

\subsubsection{Structured Pruning of MHA and FFN}\label{sec:pruning}

To compress the matrices in MHA and FFN which contribute to most of the PLM's parameters, we adopt structured pruning to compress one dimension of the matrices. As depicted in \autoref{fig:arch}, the pruning granularity of MHA and FFN are attention head and hidden dimension, respectively.

Following \cite{molchanov2017pruning}, COST-EFF has the pruning objective of minimizing the difference between pruned and original model, which is calculated by first-order Taylor expansion
\begin{equation}
    \begin{aligned}
        \left\lvert \Delta (\boldsymbol{S}) \right\rvert & = \left\lvert \mathcal{L}(\boldsymbol{X}) - \mathcal{L}(\boldsymbol{X} | \boldsymbol{h}_{i}=0, \boldsymbol{h}_{i} \in \boldsymbol{S}) \right\rvert                 \\
                                                         & = \left\lvert \sum\limits_{\boldsymbol{h}_{i}\in \boldsymbol{S}}\frac{\delta \mathcal{L}}{\delta \boldsymbol{h}_{i}} (\boldsymbol{h}_{i}-0) + R^{(1)} \right\rvert \\
                                                         & \approx \left\lvert \sum\limits_{\boldsymbol{h}_{i}\in \boldsymbol{S}}\frac{\delta \mathcal{L}}{\delta \boldsymbol{h}_{i}} \boldsymbol{h}_{i} \right\rvert
        ,
    \end{aligned}
    \label{eq:taylor}
\end{equation}
where $\boldsymbol{S}$ denotes a specific structure, i.e., a set of weights, $\mathcal{L}(\cdot)$ is the loss function and $\frac{\delta \mathcal{L}}{\delta \boldsymbol{h}_{i}}$ is the gradient of loss to weight $\boldsymbol{h}_{i}$. $\left\lvert \Delta (\boldsymbol{S}) \right\rvert$ is the importance of structure $\boldsymbol{S}$ measured by absolute value of the first-order term. For simplicity, we ignore the remainder $R^{(1)}$ in Taylor expansion.

In each Transformer layer, the structure $\boldsymbol{S}$ of MHA is the attention head while that of FFN is the hidden dimension as depicted in the lower part of \autoref{fig:arch}. Specifically, the output dimensions of $\boldsymbol{W}_Q$, $\boldsymbol{W}_K$, $\boldsymbol{W}_V$ and $\boldsymbol{W}_{FI}$ are compressed. On the contrary, the input dimensions of $\boldsymbol{W}_O$ and $\boldsymbol{W}_{FO}$ are compressed. Thus, the dimension of the hidden states remains intact in COST-EFF.
Also, as a single but drastic pruning would usually cause damage hard to recover, we use iterative pruning \cite{tan2020dropnet} in COST-EFF which gradually prunes out insignificant modules.

\subsection{Dynamic Acceleration}\label{sec:accelerate}

\subsubsection{Inference with Early Exiting}\label{sec:early-exit}

Unlike static compression, early exiting dynamically determines the computation at inference time, depending on the complexity of inputs and the perplexity of the model. Specifically, we use layer-wise early exiting, as shown in \autoref{fig:intro}, by plugging in a classifier at each Transformer layer.

Following the experimental results of ElasticBERT \cite{liu_towards_2022}, entropy-based exiting generally outperforms patience-based, we use entropy of the classifier output as the exit condition defined as
$\operatorname{H}(x) = -\sum_{i=1}^{C} \operatorname{p}(i) \ln \operatorname{p}(i)$,
where $\operatorname{p}(\cdot)$ is the probability distribution calculated by softmax function and $H(x)$ is the entropy of the probability distribution $x$. If the entropy is greater than a given threshold $H_T$, the model is hard to make a prediction at that state. Conversely, the model tends to make a certain prediction with small entropy, where the difference in the probability distribution is large and dominant.

\subsubsection{Training Multiple Exits}\label{sec:exit}

When training the model with multiple exits, the loss function of each exit is taken into account.
DeeBERT \cite{xin2020deebert} introduced a two-stage training scheme where the backbone model and exits are separately trained. However, only with the loss of the final classifier and the gradients that back-propagate, shallow layers of the backbone model are not capable of making confident predictions but rather serve the deep layers. Thus, it is necessary to introduce the loss of intermediate classifiers while training and calculating the Taylor expansion-based structure importance as \autoref{eq:taylor} in COST-EFF.

To balance the gradient from multiple classifier losses, we use gradient equilibrium following \cite{li2019improved} and scale the gradient of layer $k$ to
\begin{equation}
    \nabla_{\boldsymbol{w}_{k}}^\prime \mathcal{L} = \frac{1}{L-k+1} \sum\limits_{i=k}^{L}\nabla_{\boldsymbol{w}_{k}} \mathcal{L}_{i}
    ,
\end{equation}
where $L$ is the model depth, $\nabla_{\boldsymbol{w}_{k}} \mathcal{L}_{i}$ is the gradient propagates from layer $i$ down to layer $k$ and $\nabla_{\boldsymbol{w}_{k}}^\prime \mathcal{L}$ is the rescaled gradient.

\subsection{Collaborative Training of COST-EFF}

\subsubsection{Training with Knowledge Distillation}\label{sec:distill}

The small size and capacity of the compressed model make it hard to restore performance only with fine-tuning. Whereas knowledge distillation is used as a complement that transfers the knowledge from the original teacher model to the compressed student model. In this paper, we aim to distill the prediction and intermediate features (i.e., hidden states) as depicted in \autoref{fig:arch}.

As the inconsistency between layers is observed \cite{xin2021berxit}, simply using ground-truth labels to train a compressed multi-exit model would result in severe contradictions. Given this, we first distill the original model into a multi-exit BERT$_\text{Base}$ model with the same layers as the TA. Then, each layer output of TA is used as soft labels of the corresponding layer in COST-EFF as
\begin{equation}
    \mathcal{L}_{pred} = \sum\limits_{i=1}^{L}\operatorname{CELoss}\left(\boldsymbol{z}^{TA}_i/T, \boldsymbol{z}^{CE}_i/T\right)
    ,
    \label{eq:pred_distill}
\end{equation}
where $\boldsymbol{z}^{TA}_i$ and $\boldsymbol{z}^{CE}_i$ are the prediction outputs of TA and COST-EFF at the $i$-th layer, respectively. $T$ is the temperature factor usually set as 1.
Besides distilling the predictions, COST-EFF distills hidden states to effectively transfer the representations of TA to the student model. The hidden state outputs, denote as $\boldsymbol{H}_i (i=0, 1, \cdots, L+1)$ including embedding output $\boldsymbol{H}_0$ and each Transformer layer output, are optimized as
\begin{equation}
    \mathcal{L}_{feat} = \sum\limits_{i=0}^{L+1}\operatorname{MSELoss}\left(\boldsymbol{H}^{TA}_i, \boldsymbol{H}^{CE}_i\right)
    .
    \label{eq:repr_distill}
\end{equation}

\subsubsection{COST-EFF Procedure}\label{sec:procedure}

As mentioned in \autoref{sec:distill}, COST-EFF first distills the model into a multi-exit TA model with the same number of layers.
Specifically, we distill the predictions at this stage. Although feature distillation is typically more powerful, representations of the single-exit model are not aligned with the multi-exit model and will introduce inconsistencies during training.
Such distillation masks the trivial implementations of different PLMs to be compressed, as well as preliminarily mitigates the inconsistency between layers with a larger and more robust model. Then, the TA model is used as both the slenderization backbone and the teacher of further knowledge distillation.

During slenderization, we integrate the loss of exits into Taylor expansion-based structure importance calculation. Compared to simply using the loss of the final classifier, multi-exit loss helps calibrate the slenderization by weighing structures' contribution to each subsequent exit instead of only the final layer. In this way, the trade-off between layers can be better balanced in the slenderized model. After slenderization, the recovery training is a layer-wise knowledge transferring from TA to COST-EFF with the objective of minimizing the sum of $\mathcal{L}_{pred}$ and $\mathcal{L}_{feat}$ which mitigates the contradictions of ground-truth label training on the slenderized multi-exit model.

\section{Experimental Evaluation}

\subsection{Experiment Setup}

\paragraph{Datasets}

We use four tasks of GLUE benchmark \cite{wang2018glue}, namely
SST-2, MRPC, QNLI and MNLI. The details of these tasks are shown in \autoref{tab:dataset} and most categories of GLUE are covered.

\begin{table}[htbp]
    \centering
    \begin{tabular}{cccc}
        \toprule
        Task  & Category        & Labels & Metric \\ \midrule
        SST-2 & Single-sentence & 2      & Acc    \\
        MRPC  & Paraphrase      & 2      & F1     \\
        QNLI  & Inference       & 2      & Acc    \\
        MNLI  & Inference       & 3      & Acc    \\ \bottomrule
    \end{tabular}
    \caption{Details of the datasets.}
    \label{tab:dataset}
\end{table}

\paragraph{Comparative Methods}

We compare the following baselines and methods.
(1) Different size of BERT models, namely \textbf{BERT}$_\textbf{Base}$, \textbf{BERT}$_\textbf{6L-768H}$ and \textbf{BERT}$_\textbf{8L-256H}$, fine-tuned based on the pre-trained models of \cite{turc2019well}. (2) Representative static compression methods. \textbf{DistilBERT} \cite{sanh2019distilbert} and \textbf{TinyBERT} \cite{jiao2020tinybert}. (3) Dynamic accelerated methods. \textbf{DeeBERT} \cite{xin2020deebert}, \textbf{PABEE} \cite{zhou2020bert} and the pre-trained multi-exit model \textbf{ElasticBERT} \cite{liu_towards_2022}.

\paragraph{Model Settings}

As the number of parameters profoundly impacts the capacity and performance, we have two comparison groups with similar model sizes inside each group.
Models in the first group are with less than 20M parameters and the second group of models are of larger size above 50M parameters.
The details of model settings can be found in \autoref{tab:setting}. Notably, the results of DistilBERT are extracted from the original paper and the others are implemented by ourselves as the experiments involve different backbone models and training data. The implementation is with AdamW optimizer on a single 24GB RTX 3090 GPU, while train batch size is in \{32, 64\} and learning rate is in \{2e-5, 3e-5, 4e-5\} varying from tasks.

\begin{table}[ht]
    \setlength\tabcolsep{3.5pt}
    \centering
    \begin{tabular}{cccccc}
        \toprule
        \multirow{2}{*}{Model}    & \multicolumn{4}{c}{Size} & \multirow{2}{*}{EE}                                        \\
                                  & $L$                      & $H$                 & $A$            & $F$  &              \\ \midrule
        BERT$_\text{Base}$        & 12                       & 768                 & $12 \times 64$ & 3072 &              \\ \midrule
        BERT$_\text{8L-256H}$     & 8                        & 256                 & $4 \times 64$  & 1024 &              \\
        TinyBERT$_4$              & 4                        & 312                 & $12 \times 26$ & 1200 &              \\
        DeeBERT$_\text{12L-256H}$ & 12                       & 256                 & $4 \times 64$  & 1024 & $\checkmark$ \\
        PABEE$_\text{12L-256H}$   & 12                       & 256                 & $4 \times 64$  & 1024 & $\checkmark$ \\
        COST-EFF$_{8 \times}$     & 12                       & 768                 & $2 \times 64$  & 256  & $\checkmark$ \\ \midrule
        BERT$_\text{6L-768H}$     & 6                        & 768                 & $12 \times 64$ & 3072 &              \\
        TinyBERT$_6$              & 6                        & 768                 & $12 \times 64$ & 3072 &              \\
        DistilBERT$_6$            & 6                        & 768                 & $12 \times 64$ & 3072 &              \\
        ElasticBERT$_\text{6L}$   & 6                        & 768                 & $12 \times 64$ & 3072 & $\checkmark$ \\
        DeeBERT$_\text{12L-512H}$ & 12                       & 512                 & $8 \times 64$  & 2048 & $\checkmark$ \\
        PABEE$_\text{12L-512H}$   & 12                       & 512                 & $8 \times 64$  & 2048 & $\checkmark$ \\
        COST-EFF$_{2 \times}$     & 12                       & 768                 & $6 \times 64$  & 1536 & $\checkmark$ \\ \bottomrule
    \end{tabular}
    \caption{Settings of compressed models. $L$ is the number of layers and $H$ is the dimension of hidden states. $A$ denotes the MHA size as $head\_num \times head\_size$, and the intermediate size of FFN is $F$. Models with a check sign in the EE column adopt early exiting.}
    \label{tab:setting}
\end{table}

\subsection{Experiment Results}

\subsubsection{Overall Results}

\begin{table*}[ht]
    \setlength\tabcolsep{5pt}
    \centering
    \begin{tabular}{ccccccc}
        \toprule
        \multirow{2}{*}{Model}       & Params               & FLOPs                 & \multirow{2}{*}{SST-2} & \multirow{2}{*}{MRPC} & \multirow{2}{*}{QNLI} & \multirow{2}{*}{MNLI-m/mm}           \\
                                     & reduc.               & reduc.                &                        &                       &                       &                                      \\ \midrule
        BERT$_\text{Base}$           & 1.0$\times$          & 1.0$\times$           & 93.1                   & 90.5                  & 91.7                  & 84.4 / 84.5                          \\ \midrule
        BERT$_\text{8L-256H}$        & 7.6$\times$          & 13.5$\times$          & 88.4                   & 84.7                  & 86.6                  & 77.5 / 78.4                          \\
        TinyBERT$_4$                 & \ul{7.6$\times$}     & \ul{18.6$\times$}     & \ul{89.7}              & \ul{86.7}             & \ul{87.0}             & \ul{81.2} / \ul{81.6}                \\
        DeeBERT$_\text{12L-256H}$    & 6.0$\times$          & 14.9$\times$          & 87.5                   & 85.0                  & 86.8                  & 77.6 / 78.5                          \\
        PABEE$_\text{12L-256H}$      & 6.3$\times$          & 14.5$\times$          & 88.1                   & 85.4                  & 86.0                  & 78.6 / 78.3                          \\
        COST-EFF$_{8 \times}$ (ours) & \textbf{7.9}$\times$ & \textbf{19.0}$\times$ & \textbf{90.6}          & \textbf{87.1}         & \textbf{87.8}         & \textbf{81.3} / \textbf{81.8}        \\ \midrule
        BERT$_\text{6L-768H}$        & 1.6$\times$          & 2.0$\times$           & 91.1                   & 88.1                  & 89.6                  & 81.5 / 82.0                          \\
        DistilBERT$_6$               & 1.6$\times$          & 2.0$\times$           & 91.3                   & -                     & 89.2                  & 82.2 / ~~ - ~~                       \\
        TinyBERT$_6$                 & 1.6$\times$          & 2.0$\times$           & \ul{91.6}              & 89.2                  & \textbf{91.3}         & \textbf{84.2} / \textbf{84.4}        \\
        TinyBERT$_6$ w/o GD          & 1.6$\times$          & 2.0$\times$           & 91.2                   & 88.5                  & 90.0                  & 83.5 / 83.4                          \\
        ElasticBERT$_\text{6L}$      & 1.6$\times$          & \ul{2.4$\times$}      & 91.2                   & \ul{89.4}             & 90.5                  & 83.2 / 83.2                          \\
        DeeBERT$_\text{12L-512H}$    & 1.9$\times$          & 2.2$\times$           & 89.8                   & 89.0                  & 89.8                  & 81.8 / 82.6                          \\
        PABEE$_\text{12L-512H}$      & \ul{2.0$\times$}     & 2.2$\times$           & 89.7                   & 86.9                  & 89.2                  & 81.6 / 81.9                          \\
        COST-EFF$_{2 \times}$ (ours) & \textbf{2.0}$\times$ & \textbf{2.4}$\times$  & \textbf{92.0}          & \textbf{89.7}         & \ul{90.9}             & \ul{83.7} / \ul{83.8}                \\ \midrule
        \multicolumn{7}{c}{\textit{simple input instances}}                                                                                                                                         \\ \midrule
        TinyBERT$_4$                 & 7.6$\times$          & 18.6$\times$          & 90.1 (+0.4)            & 83.6 (-3.1)           & 87.0 (+0.0)           & 81.4 / \textbf{83.4} (+0.9)          \\
        COST-EFF$_{8 \times}$ (ours) & \textbf{7.9}$\times$ & \textbf{20.3}$\times$ & \textbf{91.5} (+0.9)   & \textbf{88.8} (+1.7)  & \textbf{87.9} (+0.1)  & \textbf{82.3} / \textbf{83.4} (+1.3) \\ \bottomrule
    \end{tabular}
    \caption{Results on GLUE development set. BERT$_\text{Base}$ is used as the baseline to evaluate the average compression and acceleration rate, i.e., Params reduc. and FLOPs reduc., which are the higher the better. TinyBERT is implemented by conducting task-specific distillation without data augmentation on the public general distilled models, while TinyBERT$_6$ w/o GD is initialized from pre-trained BERT$_\text{6L-768H}$ without general distillation. ElasticBERT$_\text{6L}$ is initialized from the first 6 layers of ElasticBERT without pooler. The best results are in bold and the second best results are underlined.}
    \label{tab:result}
\end{table*}

The results of COST-EFF and comparative methods are listed in \autoref{tab:result}. When counting parameters, we include the parameters of embeddings and use the vocabulary size of 30,522 as default. The FLOPs are evaluated by PyTorch profiler with input sequences padded or truncated to the default length of 128 tokens and are averaged by tasks.

In the first group, the models are highly compressed and accelerated, while the performance is retained at approximately 96.5\% by COST-EFF$_{8 \times}$, which is much better than the conventional pre-training and fine-tuning of BERT$_\text{8L-256H}$. Specifically, COST-EFF$_{8 \times}$ out-performs TinyBERT$_4$ in all four tasks, suggesting that a slenderized model preserving all the layers is superior to a squat one. The slenderized architecture is more likely to extract hierarchical features for hard instances while expeditiously processing simple instances.
For larger models, TinyBERT$_6$ with general distillation gains a slight advantage over COST-EFF$_{2 \times}$. Whereas COST-EFF$_{2 \times}$ has a smaller volume than TinyBERT$_6$ and does not require general distillation, the performance gap is not significant. Meanwhile, TinyBERT$_6$ without general distillation is dominated by COST-EFF$_{2 \times}$ in both efficiency and effectiveness, indicating the necessity of TinyBERT general distillation. However, a large effort is required by general distillation which pre-trains a single model of a fixed size and computation. In case the computation demand changes, pre-training yet another model can be extremely time-consuming. Compared to TinyBERT, COST-EFF has advantages in both performance and flexible inference.

\begin{figure*}[ht]
    \centering
    \includegraphics[width=0.975\textwidth]{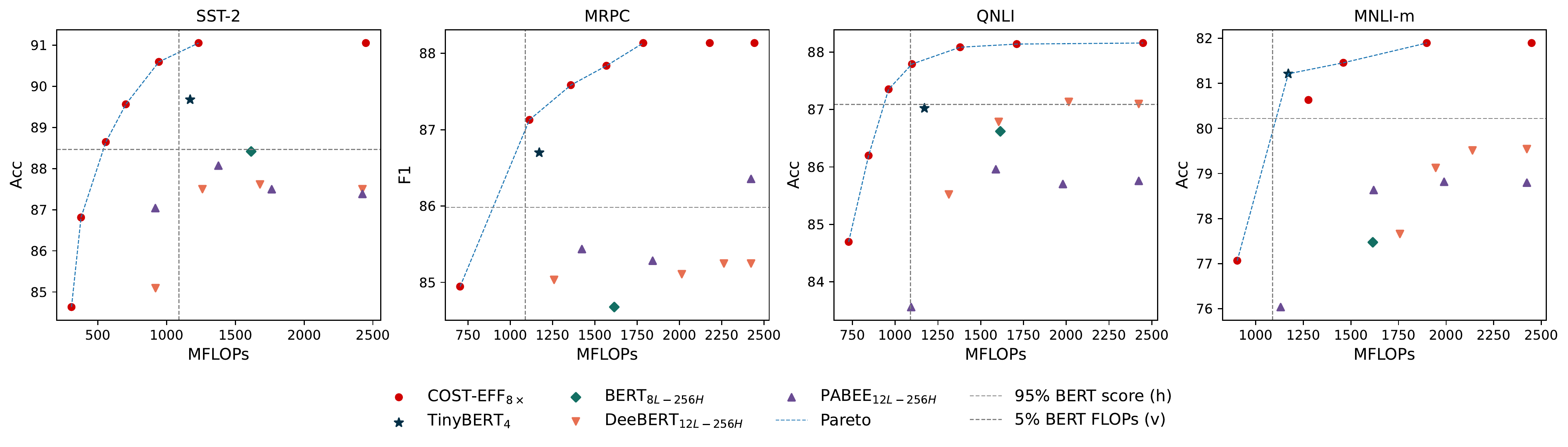}
    \caption{Performance curves of models with $8 \times$ compression rate on GLUE development set. Horizontal grey line indicates the 95\% of BERT$_\text{Base}$ performance and vertical line indicates 5\% BERT$_\text{Base}$ FLOPs.}
    \label{fig:group_1}
\end{figure*}

\begin{figure*}[ht]
    \centering
    \includegraphics[width=0.975\textwidth]{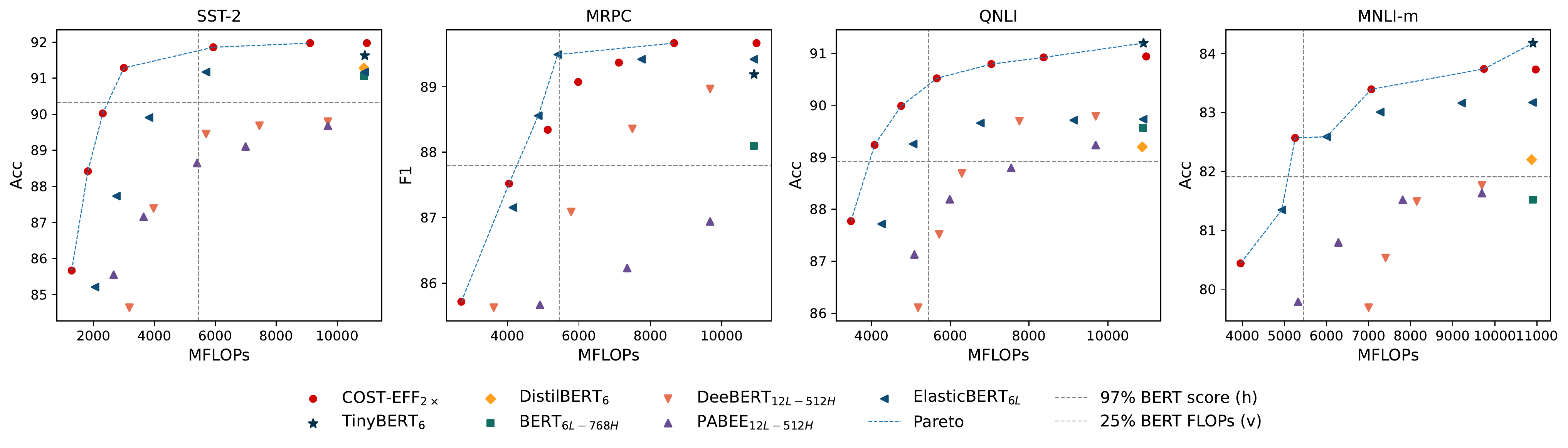}
    \caption{Performance curves of models with $2 \times$ compression rate on GLUE development set. Horizontal grey line indicates the 97\% of BERT$_\text{Base}$ performance and vertical line indicates 25\% BERT$_\text{Base}$ FLOPs.}
    \label{fig:group_2}
\end{figure*}

To demonstrate the effect of dynamic acceleration, we empirically select simple instances from the development set which are shorter (i.e., lower than the median non-padding length after tokenization). The results on simple instances exhibit extra improvements attributed to dynamic inference, which are hard to obtain with static models. Notably, shorter length does not always indicate simplicity. For entailment tasks like QNLI, shorter inputs would contain less information, which potentially aggravate the perplexity of language models. Also, we plot performance curves with respect to GLUE scores and FLOPs in \autoref{fig:group_1} and \ref{fig:group_2}. The performance curves are two-dimensional and exhibit the optimality of different methods. Aiming at obtaining the model with smaller computation and performance, we focus on the models in the upper left part of the figure, which is the Pareto frontier plotted in dashed blue lines.

As depicted in \autoref{fig:group_1} and \ref{fig:group_2}, both COST-EFF$_{8 \times}$ and COST-EFF$_{2 \times}$ outperform DistilBERT, DeeBERT, PABEE and BERT baselines. Compared with TinyBERT and ElasticBERT, COST-EFF is generally optimal. We find that early exiting reduces the upper bound of NLI performance, where both COST-EFF$_{2 \times}$ and ElasticBERT$_\text{6L}$ are inferior to TinyBERT$_6$. This issue may stem from the inconsistency between layers. Given that the complex samples in the NLI task rely on high-level semantics, the shallow layers should serve the deeper layers rather than solving the task by themselves. However, this issue does not affect global optimality. As shown in \autoref{fig:group_1}, COST-EFF$_{8 \times}$ has non-dominated performance against TinyBERT$_4$ on QNLI and MNLI, demonstrating the flexibility of our approach.

\begin{figure}[ht]
    \centering
    \includegraphics[width=0.95\columnwidth]{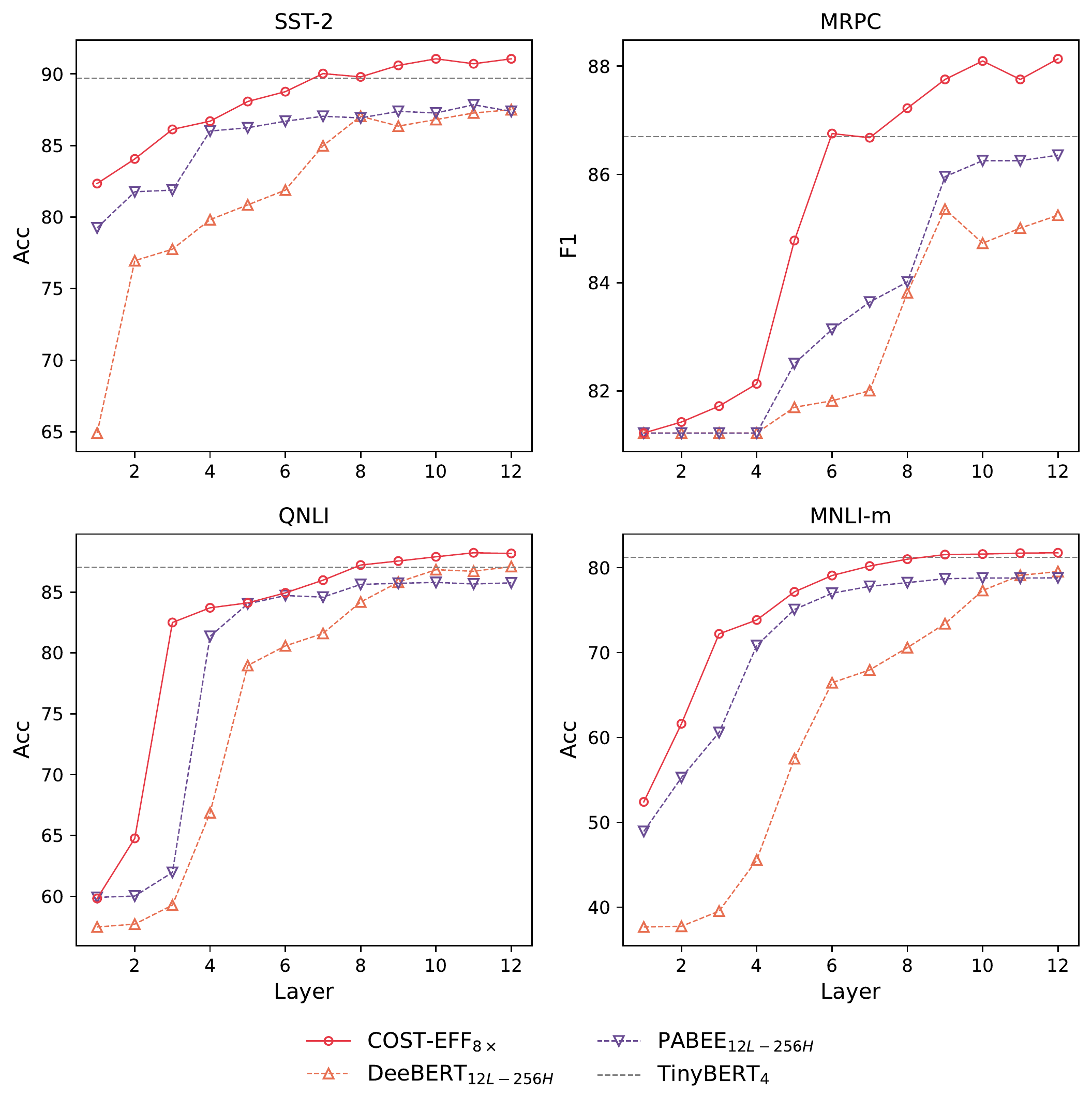}
    \caption{Layer-wise performance on GLUE development set. Horizontal lines indicate the final classifier performance of TinyBERT$_4$.}
    \label{fig:layer}
\end{figure}

The performance of models incorporating early exiting is substantially affected by each exit. In \autoref{fig:layer}, we plot the layer-wise performance of models with early exiting in the first group and the final performance of TinyBERT$_4$. COST-EFF$_{8 \times}$ achieves the dominant performance compared to DeeBERT and PABEE. Compared to TinyBERT$_4$, COST-EFF$_{8 \times}$ can achieve better performance from the 7th to 12th layer, further verifying our claim that slender models are superior to squat models in performance, benefiting from the preserved architecture and its ability to extract high-level semantics.
Another way to obtain powerful multi-exit models is alternating the backbone from BERT to the pre-trained ElasticBERT \cite{liu_towards_2022}. In view of fairness, we uniformly use BERT as the backbone of COST-EFF and comparative methods. Notably, our approach is well-adaptable to ElasticBERT and the advanced performance is exhibited in \autoref{app:elastic}.

\subsubsection{Ablation Studies}

\paragraph{Impact of knowledge distillation}

The ablation experiments of distillation strategies aim to evaluate the effectiveness of prediction and feature distillation. In this ablation study, the comparison methods are (1) ablating feature distillation and (2) alternating prediction distillation with ground-truth label training. The results shown in \autoref{tab:abl} indicate that both objectives are crucial.

\begin{table}[ht]
    \centering
    \begin{tabular}{lccccc}
        \toprule
        Model                   & SST-2 & MRPC & QNLI \\ \midrule
        COST-EFF$_{8 \times}$   & 90.6  & 87.1 & 87.8 \\
        ~~$-\mathcal{L}_{feat}$ & 87.5  & 86.8 & 86.4 \\
        ~~$-\mathcal{L}_{pred}$ & 88.6  & 82.4 & 84.2 \\ \bottomrule
    \end{tabular}
    \caption{Ablation results on GLUE development set with $8 \times$ compression. Feature distillation is ablated in $-\mathcal{L}_{feat}$, while ground-truth label is used to replace prediction distillation in $-\mathcal{L}_{pred}$. FLOPs of two ablated methods are ensured more than COST-EFF$_{8 \times}$.}
    \label{tab:abl}
\end{table}

Attributing to the imitation of hidden representations, COST-EFF$_{8 \times}$ has an advantage of 1.6\% in performance compared to training without feature distillation.
Without prediction distillation, the performance drops more than 3.4\%. Previous works of static compression, e.g., TinyBERT \cite{jiao2020tinybert} and CoFi \cite{xia2022structured}, are generally not sensitive to prediction distillation in GLUE tasks, as the output distribution of the single-exit teacher model is generally in accordance with the ground-truth label. However, a large decrease in COST-EFF performance is observed in \autoref{tab:abl} if prediction distribution is ablated. The result indicates that pursuing the ground truth at shallow layers can deteriorate the performance of deep layers. Such inconsistency between shallow and deep layers commonly exists in early exiting models, which is particularly hard to balance by compressed models with small capacity. Instead, COST-EFF introduces an uncompressed TA model to mitigate the contradiction at an early stage and transfer the balance through prediction distillation.

\paragraph{Impact of collaborative training}

\begin{figure}[ht]
    \centering
    \includegraphics[width=0.95\columnwidth]{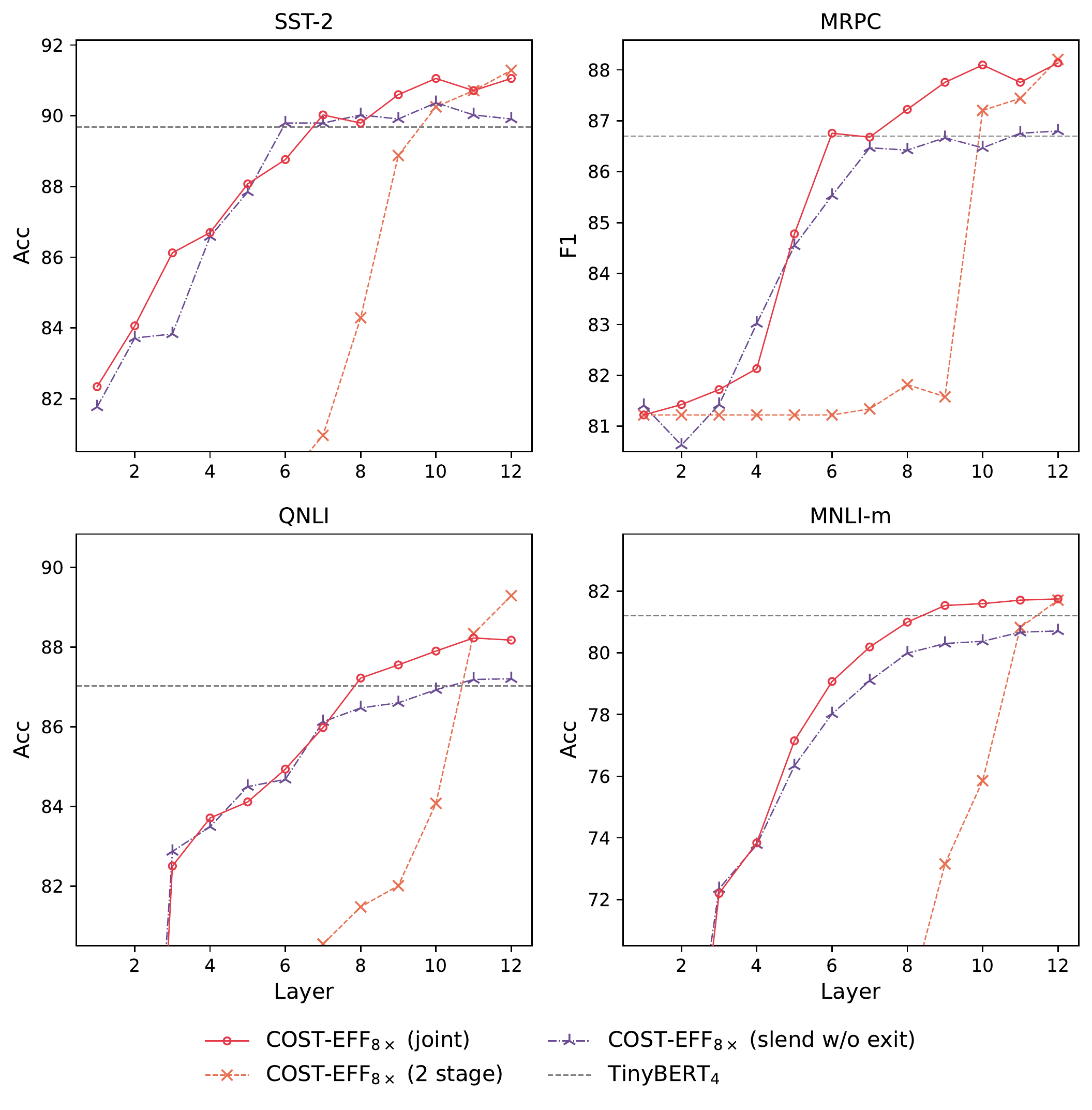}
    \caption{Layer-wise performance (zoomed) of collaborative training ablation study on GLUE development set. Horizontal lines indicate the final classifier performance of TinyBERT$_4$. COST-EFF$_{8 \times}$ (slend w/o exit) is slenderized without the calibration of exits.}
    \label{fig:abl}
\end{figure}

In this paper, we propose a collaborative approach for model slenderization and exit training, intended to calibrate the pruning of shallow modules. To validate the effectiveness of the training strategy, we ablate the collaborative training at different times. First, we implemented a two-stage training mode as DeeBERT does. Also, we implement COST-EFF$_{8 \times}$ with exit loss ablated before and during slenderization. The layer-wise comparison of the above methods is shown in \autoref{fig:abl}.

Intuitively, two-stage training has an advantage on the final layer over collaborative training, as the inconsistency between layers is not introduced. However, the advantage diminishes in shallow layers, leaving the general performance unacceptable. Compared to slenderizing without exit loss, our approach has an advantage of 1.1\% to 2.3\%. Notably, slenderizing without calibration of exit can still achieve similar performance to COST-EFF at shallow layers, suggesting that the distillation-based training is effective in restoring performance. However, the inferior performance of deep layers indicates that the trade-off between layers is not well-balanced, since the slenderization is conducted aiming at optimizing the performance of the final classifier.

\section{Conclusion}

In this paper, we statically slenderize and dynamically accelerate PLMs in the pursuit of inference efficiency as well as preserving the capacity. To integrate the perspectives, we propose a collaborative optimization approach that achieves a mutual gain of static slenderization and dynamic acceleration. Specifically, the size of PLM is reduced in model width, and the inference is adaptable to the complexity of inputs without introducing redundancy for simple inputs and inadequacy for hard inputs. Comparative experiments are conducted on GLUE benchmark and verify the Pareto optimality of our approach at high compression and acceleration rate.

%% file: appendix.tex
\section{ElasticBERT as Backbone}\label{app:elastic}

\begin{figure*}[ht]
    \centering
    \includegraphics[width=0.95\textwidth]{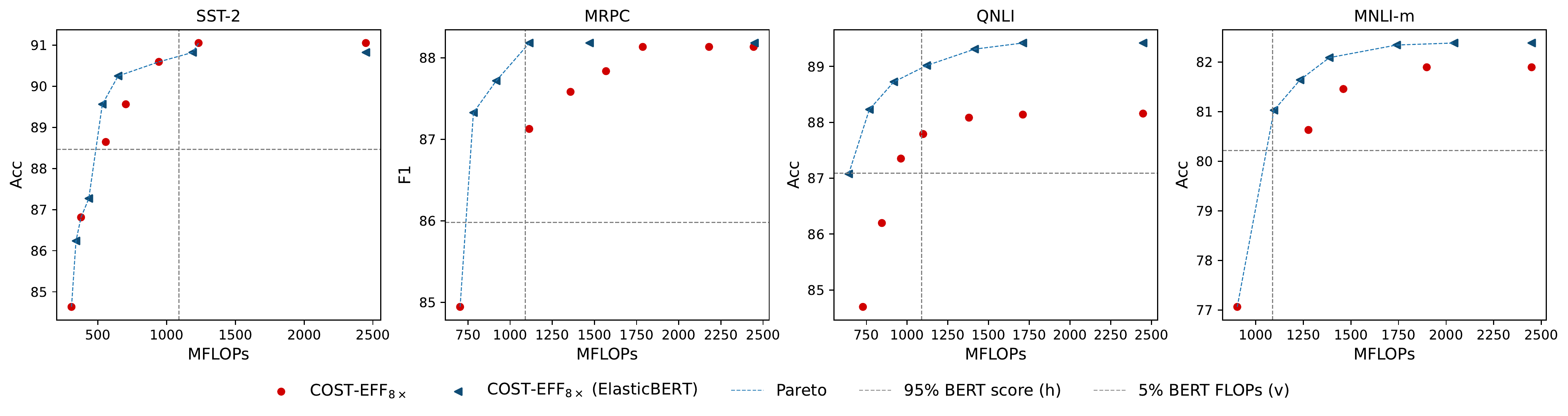}
    \caption{Performance curves COST-EFF on GLUE development set with BERT and ElasticBERT. Horizontal grey line indicates the 95\% of BERT$_\text{Base}$ performance and vertical line indicates 5\% BERT$_\text{Base}$ FLOPs.}
    \label{fig:group_1_elastic}
\end{figure*}

\begin{figure}[ht]
    \centering
    \includegraphics[width=0.95\columnwidth]{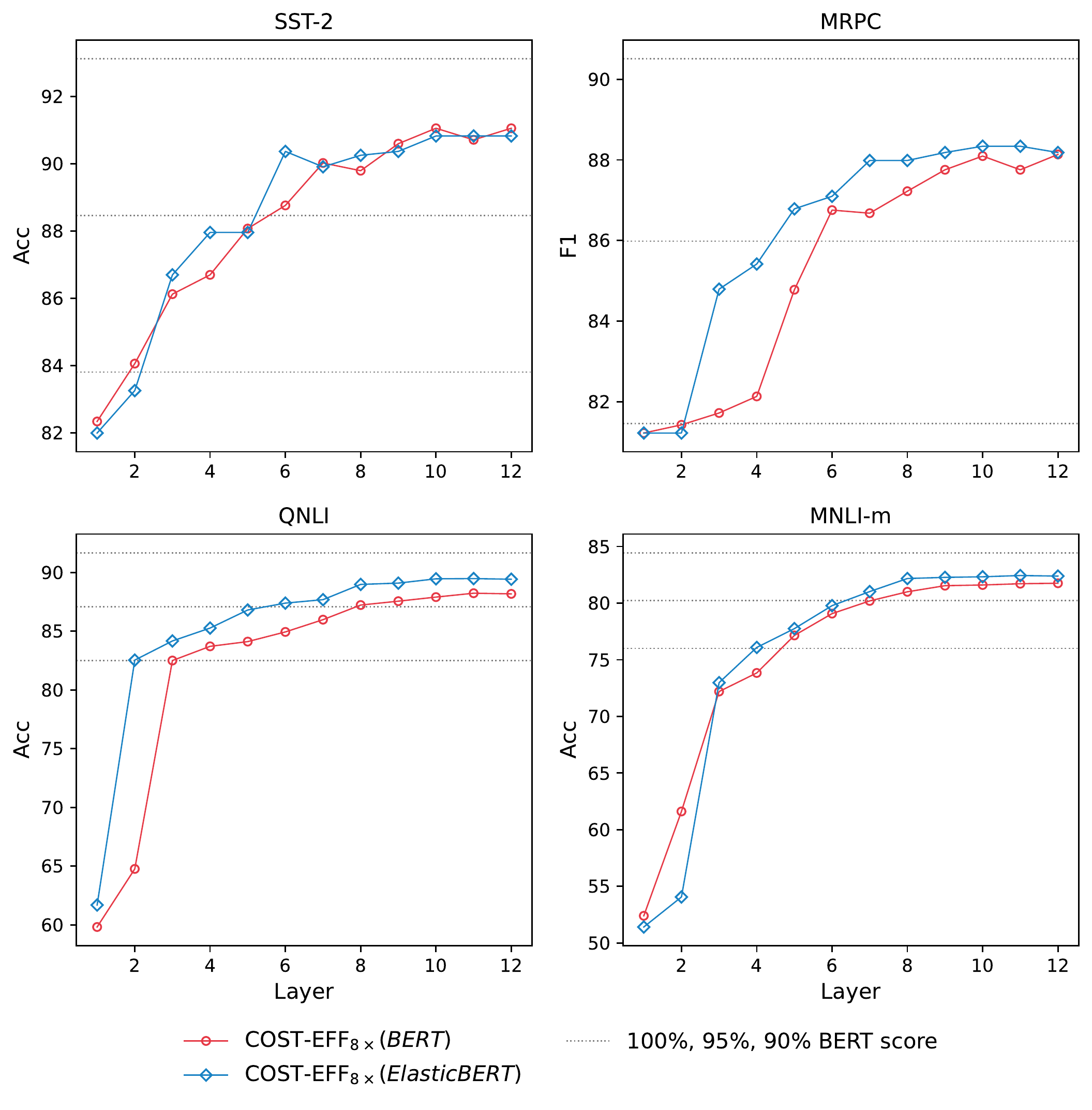}
    \caption{Layer-wise performance on GLUE development set. Horizontal lines indicate 100\%, 95\% and 90\% of BERT$_\text{Base}$ performance from top to bottom.}
    \label{fig:layer_elastic}
\end{figure}

We implemented COST-EFF with the backbone of ElasticBERT and obtain better performance than BERT backbone. The global results are listed in \autoref{tab:result_elastic}. Also, we plot performance curves and layer-wise performance in \autoref{fig:group_1_elastic} and \autoref{fig:layer_elastic}, respectively.

\begin{table*}[ht]
    \setlength\tabcolsep{5pt}
    \centering
    \begin{tabular}{ccccccc}
        \toprule
        \multirow{2}{*}{Model}              & Params      & FLOPs                 & \multirow{2}{*}{SST-2} & \multirow{2}{*}{MRPC} & \multirow{2}{*}{QNLI} & \multirow{2}{*}{MNLI-m/mm}    \\
                                            & reduc.      & reduc.                &                        &                       &                       &                               \\ \midrule
        BERT$_\text{Base}$                  & 1.0$\times$ & 1.0$\times$           & 93.1                   & 90.5                  & 91.7                  & 84.4 / 84.5                   \\ \midrule
        COST-EFF$_{8 \times}$ (BERT)        & 7.9$\times$ & 19.0$\times$          & 90.6                   & 87.1                  & 87.8                  & 81.3 / 81.8                   \\
        COST-EFF$_{8 \times}$ (ElasticBERT) & 7.9$\times$ & \textbf{19.1}$\times$ & \textbf{90.8}          & \textbf{88.1}         & \textbf{89.0}         & \textbf{81.6} / \textbf{82.3} \\ \bottomrule
    \end{tabular}
    \caption{Results of COST-EFF on GLUE development set with BERT and ElasticBERT as the backbone.}
    \label{tab:result_elastic}
\end{table*}

